\documentclass[10pt,twocolumn,letterpaper]{article}

\usepackage{cvpr}
\usepackage{times}
\usepackage{epsfig}
\usepackage{graphicx}
\usepackage{amsmath}
\usepackage{amssymb}
\usepackage{comment}
\usepackage{caption}
\usepackage{subcaption}

\usepackage[pagebackref=true,breaklinks=true,letterpaper=true,colorlinks,bookmarks=false]{hyperref}

\cvprfinalcopy 

\def\cvprPaperID{****} 

\ifcvprfinal\fi
\begin{document}

\title{Unsupervised Multi-label Dataset Generation from Web Data}

\author{Carlos Roig \qquad \qquad David Varas \qquad  \qquad Issey Masuda \\
\\
Juan Carlos Riveiro \qquad Elisenda Bou-Balust \\
\\
Vilynx Inc.\\
{\tt\small {\{carlos,david.varas,issey,eli\}@vilynx.com}}}
\maketitle

\begin{abstract}
This paper presents a system towards the generation of multi-label datasets from web data in an unsupervised manner. To achieve this objective, this work comprises two main contributions, namely: a) the generation of a low-noise unsupervised single-label dataset from web-data, and b) the augmentation of labels in such dataset (from single label to multi label).
The generation of a single-label dataset uses an unsupervised noise reduction phase (clustering and selection of clusters using anchors) obtaining a 85\% of correctly labeled images. An unsupervised label augmentation process is then performed to assign new labels to the images in the dataset using the class activation maps and the uncertainty associated with each class. This process is applied to the dataset generated in this paper and a public dataset (Places365) achieving a 9.5\% and 27\% of extra labels in each dataset respectively, therefore  demonstrating that the presented system can robustly enrich the initial dataset.
\end{abstract}

\section{Introduction}
Deep learning methods are widely used for many computer vision tasks. Among these tasks, image classification -which consists on assigning a label to each image- is one of the most researched by the community due to their myriad applications. To perform this task, a dataset consisting of images and their corresponding labels is required. Common datasets for image classification are  MNIST (handwritten digits) \cite{MNIST}, ImageNet (object categories) \cite{imagenet_cvpr09} (e.g. ostrich, kite or snail), Places365 (scene categories) \cite{Places365} (e.g. bedroom, coast or downtown) or LFW (person identities) \cite{LFW}.


The aforementioned datasets have one thing in common: they are manually generated. Curating a dataset is a very expensive and time-consuming task due to a) the number of samples required for classification using Deep Learning (e.g. ImageNet consists on 1.6M images) and b) the necessity to involve field experts to manually (or semi-automatically) decide which label corresponds to each sample. Because of this, the number of datasets available is limited. While the available datasets might be enough for research, their size and lack of diversity is currently precluding the growth of industrial applications. This has caused that nowadays industrial image classification is limited to big corporations who can afford the generation of such datasets.

\begin{figure}[t]
\begin{center}
   \includegraphics[width=1\linewidth]{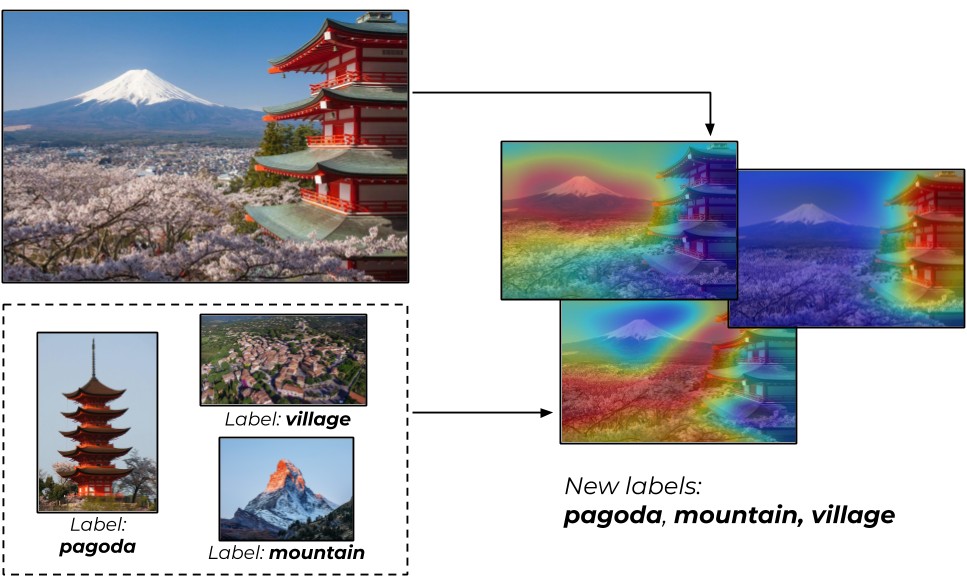}
\end{center}
\vspace{-6pt}
   \caption{The label augmentation pipeline presented in this work is an unsupervised approach for generating multiple labels using a single labeled dataset. A convolutional neural network is used to localize salient regions in the image, their associated probability scores and class uncertainty to accurately predict multiple labels for an image.}
\label{fig:first_image}
\end{figure}

Unsupervised dataset generation from Web Data is foreseen as a solution to mitigate this problem. By using image search engines, it is possible to gather samples that correspond to a text query. Moreover, the images can include metadata information which can be used to retrieve the most relevant samples for a given task. This allows the generation of datasets in a faster and more cost-effective way\cite{Webvision}. However, due to the nature of web data, such datasets usually contain a lot of noisy samples. If the input query is not specific enough or the number of required samples is too large, the amount of noisy samples could become exponentially larger.


The noisy samples in unsupervised datasets can be divided into two groups: 1) incorrectly labeled images (images with a wrong label) and 2) images partially labeled (images with a correct label, but lacking other labels that could also have). This paper presents a novel approach to mitigate both types of noise.


Many approaches have been proposed in the literature to deal with the problem of incorrectly labeled/noisy samples \cite{KrauseSHZTDPL15, sun2018learning}. However, these approaches rely on the images having attached metadata, which is not always available. In this work we present a method to generate an unsupervised dataset with noise reduction using only the ranking of the samples coming from their corresponding web searches.

To mitigate the second type of noise, the incomplete labeling of images, some approaches have been found in the literature \cite{Durand2019}. However, this problem still remains unsolved, because the previous approach requires to iterate multiple times over the dataset to learn new missing labels. 

This paper proposes a novel system that is able to output multiple labels from a single-label dataset in an unsupervised manner (Figure \ref{fig:first_image}), therefore providing label augmentation for samples that are incompletely labeled (single-labeled images).

Our contribution in this paper is twofold: 
\begin{enumerate}
    \item A model to unsupervisedly generate a single-label dataset using web data with a noise reduction mechanism.
    \item A model to provide unsupervised label augmentation, to mitigate the incomplete-labeling noise.
\end{enumerate}

This paper is structured as follow: Section \ref{sec:Related} reviews previous work in unsupervised dataset generation and noise reduction, emphasizing the differentiation between previous work and the main contributions presented in this paper. Section \ref{sec:system_overview} analyzes the proposed solution and system overview. In section \ref{sec:implementation_details}, the implementation details of such system are described to encourage reproducibility. Quantitative and qualitative results obtained from this work (illustrated with both Places365 and our custom dataset) are presented in Section \ref{sec:results}. Finally, Section \ref{sec:conclusions} concludes with a review of the work done, the insights obtained during the development of this work and future lines of research.

\section{Related work}

    
\label{sec:Related}
The work presented in this paper involves three areas of research:  a) Dataset generation from web data (and different approaches to deal with incorrectly labeled samples), b) learning from noisy labeled data and (c) label augmentation (learning multiple classes from single labeled images to deal with incomplete labeling noise).

{\bf Dataset generation from web data} is a very relevant field of research due to the cost of manually generated datasets from scratch. Moreover, thanks to the huge amount of partially annotated data that can be gathered from the web, the applications of web generated datasets are ubiquitous. Many of the most relevant datasets in computer vision have been generated by harvesting information from the web: Image datasets (ImageNet \cite{imagenet_cvpr09}, Places365 \cite{Places365}, SUN397 \cite{SUN397}, PASCAL VOC \cite{PASCALVOC}, MS COCO \cite{MSCOCO} or Open Images \cite{OpenImages}), video datasets (Youtube-8M \cite{YouTube-8M}, ActivityNet \cite{activitynet}, AVA \cite{AVA} or Kinetics \cite{Kinetics}) and audio event recognition datasets (AudioSet \cite{audioset}). The primary data source for all the aforementioned datasets is web data. However, all of them have been manually annotated and cleaned to some extent (are not unsupervised).

Other datasets generated from web data are completely unsupervised, such as WebVision \cite{Webvision}, consisting on 5 thousand visual concepts mined from public sources reaching a total of 16 million noisy images. This dataset is gathered by using textual search queries in image search engines. WebVision, while extremely valuable because of the number of samples and the minimum supervision required, has still some noise.
TourTheWorld \cite{TourTheWorld} is also generated in an unsupervised manner. In this dataset, labels are generated for landmark images based on their location metadata. This is done by creating clusters of spatially nearby pictures and then clustering the visual features of each sample for pruning purposes. However, while this process relieves the need of manually annotating the samples, it does not provide name labels corresponding to each cluster of nearby samples. Therefore, the problem of autonomous generation of unsupervised datasets with named classes remains still unsolved.

 Previous work from the authors,  WorldLocations \cite{WorldLocations}, addresses this challenge by introducing a step of unsupervised classname-annotation and noise reduction. This resulted in a landmarks dataset with named labels that did not require any human supervision or cleaning. The fist contribution of this work tackles the generalization of WorldLocations for its usage with other types of web data.

\begin{figure*}[ht]
\begin{center}
\includegraphics[width=\linewidth]{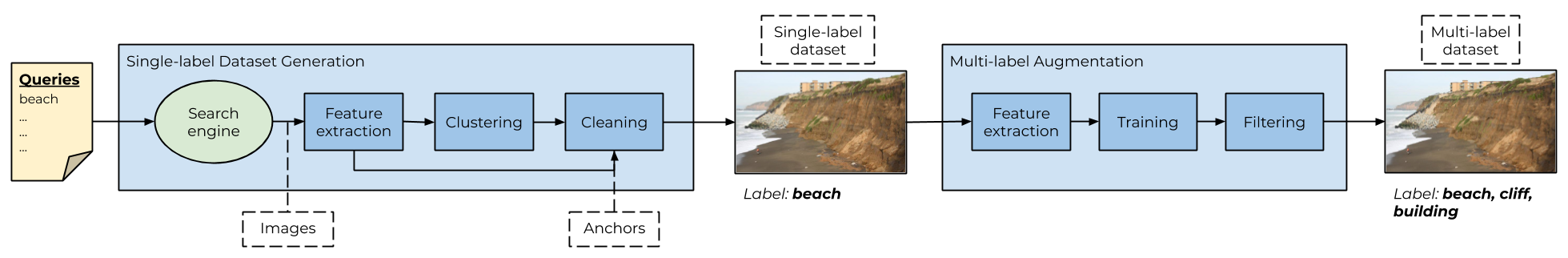}
\end{center}
\vspace{-6pt}
   \caption{Proposed system. The first part of the process, Single-label Dataset Generation, uses search engines to find images based on queries associated with the dataset labels. Metadata associated with each image is used to select a set of anchors and to eliminate noisy samples. This procedure results in a single-label dataset.  Multi-label augmentation is then performed using this dataset. In the process, a classification layer is trained to do multi-class predictions, by using features extracted from the image, class scores and uncertainties.}
\label{fig:system_overview}
\end{figure*}

Besides the generation of new datasets in unsupervised ways, many works have focused on {\bf learning discriminating visual features based on web data} using deep learning techniques. Chen and Gupta \cite{ChenG15a} crawled images from thousands of labels in order to train a CNN for object detection and localization that matched the results of a network trained with ImageNet. In \cite{LearningfromWebData}, Sun {\it et al.} studied the differences between web and standard datasets and then used web data to assist the training of networks for image classification and localization. Krause {\it et al.} \cite{KrauseSHZTDPL15} outperformed several fine-grained recognition task by increasing the training samples using noisy web data. Similarly, Joulin {\it et al.} \cite{JoulinMJV15} used captioned images in order to learn text relationships through the visual features of the images mined. All these works prove that web data can be used to train models that perform well compared to models trained only with hand-crafted datasets.

Moreover, more recent works analyze the usage of noisy data to assist the training of many vision tasks. Sun {\it et al.} \cite{SunSSG17} used the JFT-300M dataset to evaluate the performance of current vision tasks, leading to state-of-the-art results in image classification, object detection, semantic segmentation and pose estimation. Guo {\it et al.} \cite{CurriculumNet} used the idea of curriculum learning for their CurriculumNet. By measuring data complexity using cluster densities, they were able to implement a training schedule that led them to achieve state-of-the-art results in WebVision, Imagenet, Clothing-1M \cite{Clothing1M} and Food-101 \cite{Food101}. In this context, this work addresses the training of a neural network using an unsupervised dataset generated from web data.

{\bf Multi-Label Image Recognition} is one of the pivotal challenges in computer vision, because of the multi-label nature of real-world images (these images always contain different labels to be recognized). In order to address this problem, Wang {\it et al.} \cite{Wang2017} proposed using an LSTM sub-network to generate new regions within an image and the relations between these regions. In \cite{Zhu2017}, Zhu {\it et al.} exploited semantic and spatial information using a deep neural network that captured the relations between them, allowing them to predict multiple labels. Ge {\it et al.} \cite{Ge2018} presented a weakly supervised approach to curriculum learning, that first obtains intermediate object localizations and pixel labels to then perform task-specific trainings. The previous works relies on attention mechanisms and region proposals to identify new labels. All the aforementioned works rely on poor baselines, which makes it very difficult to compare the different proposed solutions. Wang {\it et al.}. In \cite{QWang2018}  Liu {\it et al.} addressed this issue by studying data augmentation techniques and ensembles for multi-label classification, used to review three datasets for benchmarking. This work used a weakly supervised detection model that was used to distill information for training the classification network. Finally, Durand {\it et al.} \cite{Durand2019} proposed a new classification loss that uses the amount of known labels for learning with partially labeled datasets in an iterative way. Even though the research community is very active in multi-label image classification, there is a lack of major multi-label image datasets for solving these problems. Moreover, to the author's knowledge there has not been any multi-label  dataset generated in an unsupervised manner. To this end, the work presented in this paper addresses the unsupervised generation of a multi-label image dataset through label augmentation.

\section{System Overview}
\label{sec:system_overview}
In this section, the proposed system for multi-label dataset generation is presented.
The approach described in this paper is composed by two main modules: First, a semi-supervised single-label dataset generation system that, given a set of queries, extracts images from web data and assigns them a label. The output of this system (a single-label dataset) is the input of the second module, which aims to generate multiple labels for images in an unsupervised manner.
The complete pipeline is illustrated in Figure \ref{fig:system_overview}.

\subsection{Single-label dataset generation from web data}
\label{subsec:single_label_gen}

\begin{figure*}[ht]
\begin{center}
\includegraphics[width=\linewidth]{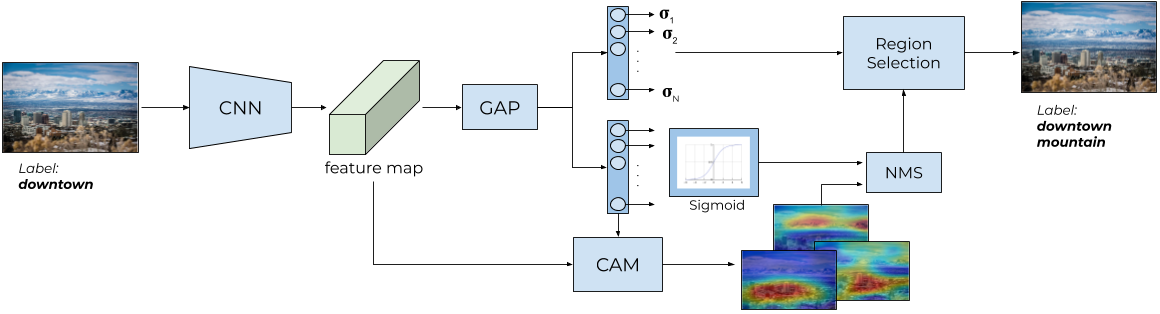}
\end{center}
   \caption{Architecture of the multi-label prediction system. First, a feature map is extracted from the image using a neural network. This map is used to compute the label scores, class uncertainties and Class Activation Maps (CAMs). Then, multiple regions are proposed using the scores and CAMs. Finally, class uncertainty is used to select multiple labels.}
\label{fig:label_augment_pipeline}
\end{figure*}

The pipeline starts with the set of $K$ labels that represent the classes of the desired dataset. These labels are queried into an image search engine. Due to the noisy nature of the resulting images, a cleaning step is further required before using these images to train a classifier. In order to perform this process, an initial seed consisting on a set of anchors is generated. These anchors correspond to the first retrieved images in the aforementioned queries as they usually correspond to the most reliable results. These anchors are later used in the cluster selection step.

Then, a feature extractor module is used to obtain a representation of the downloaded images. Any kind of feature extractor could be used as long as it is capable to provide a reliable semantic representation of the image. In the proposed system, a ResNet-50 trained with ImageNet is used for this purpose.

The feature extraction step is followed by a clustering which aims to create an over-segmentation of the data in order to select those clusters that better represent the desired class. The features of the anchor images are not considered in the clustering process. Finally, a cluster selection step is performed computing the distances between the anchors and the centroids of the clusters. The clustering is proposed in contrast of an exhaustive comparison feature-to-anchor as it helps to quantize the feature space and thus reduce the noise added by spurious. This selection is key for reducing the noise in the dataset.

\subsection{Unsupervised label augmentation}
\label{subsec:label_augmentation}

Datasets for image classification tasks are usually annotated with the predominant class in the image \cite{ imagenet_cvpr09, Places365, Webvision}. However, in real images, more than one class from the dataset may be visible in the scene. Two main problems arise in this situation: First, the model penalizes the prediction of correct visible labels during training when these labels are not the predominant class in the scene. Second, datasets containing classes which are hyperonyms of other classes (i.e. {\it stadium} and {\it football stadium}) only consider as a correct label a single semantic level while information at different levels could be exploited.

These two problems appear due to the fact that information associated with classes in the dataset is not fully exploited for training. Despite having the visual representation of several classes in images that do not contain their label, they have not been annotated with multiple labels due to the high cost of this process. 

In this section, the generation of new labels for annotated images is presented (Figure \ref{fig:label_augment_pipeline}). Although the objective of this work is to generate a multi-label annotation from single labeled images from web data, this process may be extended to any number of input and output labels using the same technique. New labels assigned to images are transferred from existing classes in the dataset. Moreover, when new classes are added to the dataset, their labels may be easily transferred to the previous dataset images using this process.

First, a convolutional neural network is trained for the task of single label classification using the initial dataset. A single classifier per class is trained for this purpose using binary cross entropy loss $\mathcal{L}^{CE}$ with sigmoid as activation function. Following \cite{cvpr2016_zhou}, we introduce a term $\sigma_k$ in the loss to model the uncertainty of the prediction for the $k_{th}$ class:
\begin{equation}\label{eq:loss}
    L = \sum_k \frac{1}{2 \sigma^{2}_{k}} \mathcal{L}^{CE}_{k} + \log \sigma_{k}
\end{equation}
Using this loss function, the model learns to predict the \textit{Heteroscedastic Uncertainty} for each class. This type of uncertainty depends on the input data and is predicted as a deterministic mapping from inputs to model outputs. As it can be observed, when the model predicts something wrong, it learns to attenuate the residual term by increasing the uncertainty of the class $\sigma^{2}_k$. This information is further used to discard predictions with large uncertainty for the model.  

The backbone of the network is used as a feature extractor. Then, a Global Average Pooling (GAP) and a classification layer with sigmoid activations $W_s$ that outputs the number of desired classes are stacked on top of the backbone. In this work, GAP is used as its loss encourages the network to identify the extent of the objects in the image in contrast with Global Max Pooling (GMP), which encourages the system to identify just one discriminative part \cite{cvpr2016_zhou}. This is because the average of the feature map is performed, so the maximum can be obtained by finding all the discriminative parts of the class because all low activations reduce the output of a particular feature map. On the other hand, low scores for all image regions except the most discriminative one do not impact the score in the GMP.

In order to assign new labels to a given image, several considerations should be taken into account. First, the class should be visible with a minimum degree of quality. In other words, the area of the class must be larger than a certain threshold. Second, a single class may be assigned to each region of the image. As a consequence, overlapping classes should be suppressed. Third, the model should be confident about its prediction to give a new label to an image.

Taking this into account, for each image in the dataset, the Class Activation Map (\textit{CAM}) of all the classes are used to generate new labels. The score associated with a given class $k$ is computed as:
\begin{equation}
    S_k = \sum_i w^{k}_i \sum_{x,y}f_i(x,y) = \sum_{x,y} \sum_i w^{k}_i f_i(x,y)
\end{equation}
where $f_i$ is the $i_{th}$ slice of the feature volume map extracted from the last convolutional layer of the model and $w^{k}_{i}$ is the $i_{th}$ weight of the classification layer of class $k$. Then, the CAM associated with this class is defined as:

\begin{equation}
    CAM_k(x,y) = \sum_i w^k_i f_i(x,y)
\end{equation}

Each position of the CAM represents the pixel probability of belonging to a given class. Then, the $k_{th}$ class is localized in the image by finding those pixels of $CAM_k$ with a probability larger than $T_p$. 

A Non Maximum Supression (NMS) step is adopted on the regions with large probability extracted from the CAMs. In the case of two or more regions sharing an area larger than a given IoU threshold $T_{IoU}$, the region belonging to the class with larger score $S_k$ is kept. Labels with at least one non suppressed region after this process are assigned to the image under analysis.
\begin{table}[ht]
    \centering
    \begin{tabular}{r||c|c}
        & Precision & Labels added \\
        \hline
        Custom dataset & 83.5\% & 9.5\% \\
        Places365 & 81\% & 27\%
    \end{tabular}
    \caption{Results of the multi-label stage. Precision of the new labels has been measured together with the number of new labels added}
    \label{tab:results_multi}
\end{table}

\section{Implementation details}
\label{sec:implementation_details}
In this section, several details related with the implementation of the label augmentation pipeline are presented in order to encourage reproducibility.

The backbone network used in the feature extraction step in Section \ref{subsec:label_augmentation} is a ResNet-50 trained with Places365. The feature maps are extracted from the last layer of this network before classification. These maps have a shape of 7x7x2048, due to the fixed input image size of 224x224.

The feature map is resized using a bilinear interpolation. Ideally, this interpolation should be performed to fit the original size of the image. As the processing time of this operation grows exponentially with the size of the resulting map, it is performed in two steps. First, the feature map is resized to 63x63x2048, from which CAMs are extracted. Then, CAMs are resized to the input image shape.

The classification and uncertainty layers are trained following the loss in Equation \ref{eq:loss}. These layers have output size  96 and 365 for our custom dataset and Places365 respectively. The training process uses the same data augmentation techniques that are used in \cite{Places365}.

\section{Results}
\label{sec:results}
In this section, a set of experiments is presented in order to validate the proposed method. These experiments have been conducted using the settings presented in Section \ref{sec:implementation_details}. 

First, the Single-label Dataset Generation process is assessed. To this end, a single-label dataset has been generated using the aforementioned pipeline for evaluation purposes. This dataset is composed by 250K images that belong to 96 classes associated with different types of scenes and places. The quality of the automatically generated single-label dataset has been evaluated selecting a random subset of the images and manually assessing whether the label associated with each image belongs to the requested class or not. The accuracy given by such pipeline is 85\%.
\begin{figure}[t]
\begin{center}
\includegraphics[width=1\linewidth]{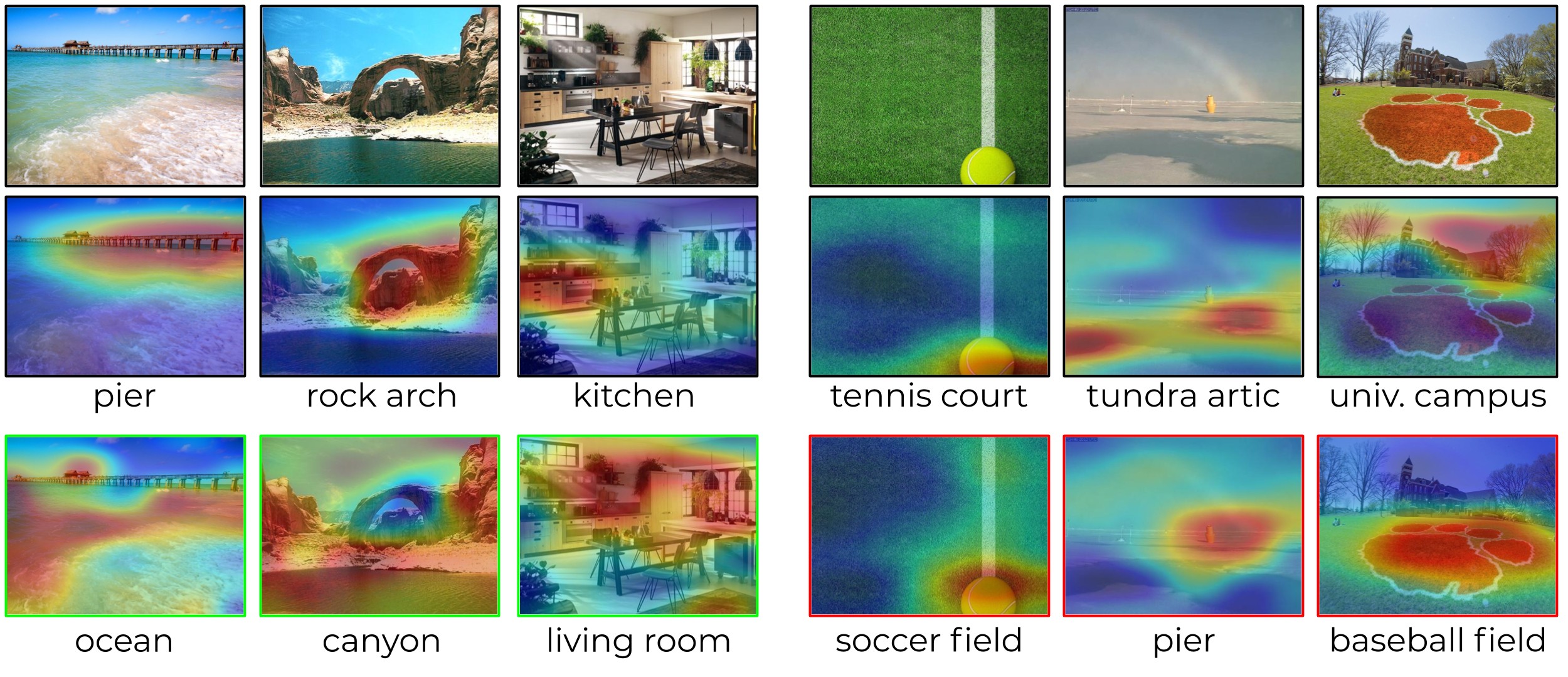}
\end{center}
\vspace{-10pt}
   \caption{Examples of multi-labeled images of our custom dataset. The first row shows the original image. Second row shows the CAMs for the original label whereas third row shows correct (\color{green}green\color{black}) and incorrect (\color{red}red\color{black}) generated labels with their CAMs}
\label{fig:results_seal}
\end{figure}

In order to assess the Multi-label Data Augmentation process, two datasets have been used. The first dataset is created using the Single-label Dataset Generation described before. Also, we used Places365 \cite{zhou2014learning}, which is a publicly available single-label dataset that contains several images with more than one visible class. This dataset has been selected to allow further comparison with other methods. Both datasets have been evaluated taking a random subset of the predicted new labels and manually assessing them. The results can be found in Table \ref{tab:results_multi}.

As it can be observed, the proposed system increases the number of labels with a large precision. This increase depends on the nature of the dataset. The results of Table \ref{tab:results_multi} show that the number of labels added for the Places365 dataset is larger than the labels generated for the custom dataset. Visually inspecting the dataset leads to an understanding that Places365 contains more images where a class-collision happens, meaning that more than one class should have been annotated but due to the intrinsic single-label nature of the dataset, only one was taken into account. In contrast, in the custom dataset, as the number of classes is smaller than in Places365, fewer classes collide and thus the number of labels added is not as big as in the later.

\subsection{Qualitative results}
\label{subsec:qualitative_results}
In Figure \ref{fig:results_seal}, some example images from our custom dataset and their corresponding labels are shown. For each image, the CAMs of the annotated and generated labels are presented. Moreover, some examples of correct and incorrect label augmentation are shown. As it can be observed, the highlighted regions of the correct examples focus on regions that visually represent the generated label. The incorrect examples show views that are not common for the different classes or patterns that are visually similar to other classes from the dataset (e.g. the paw in the field of the {\it university campus} has colors and lines that are mistaken with a {\it baseball field}).

\begin{figure}[t]
\begin{center}
\includegraphics[width=1\linewidth]{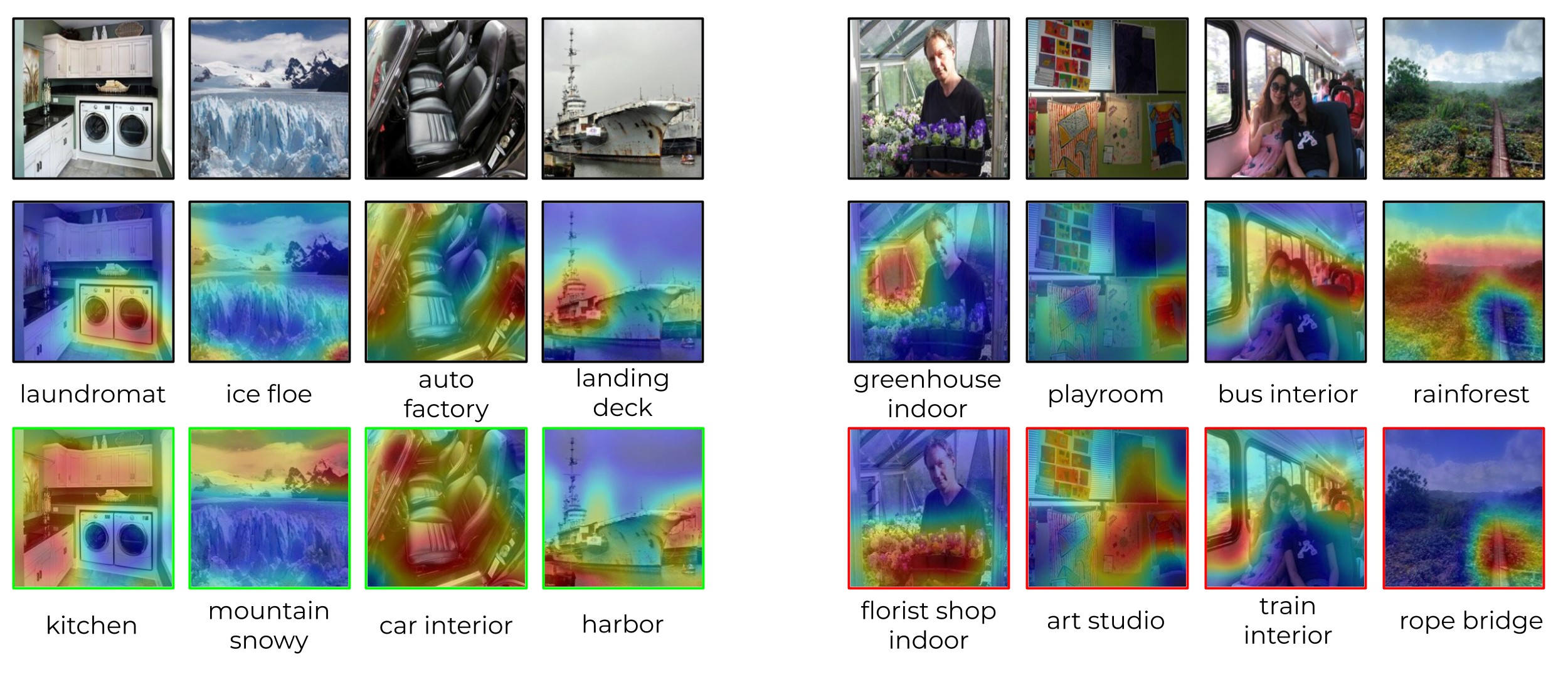}
\end{center}
\vspace{-10pt}
   \caption{Examples of the Places365 dataset. The first row shows the original image. Second row shows the CAMs for the original label whereas third row shows correct (\color{green}green\color{black}) and incorrect (\color{red}red\color{black}) generated labels with their CAMs}
\label{fig:results_places}
\end{figure}

Several examples showing images from Places365 dataset can be found in Figure \ref{fig:results_places}. The correct examples show similar patterns to the examples presented in Figure \ref{fig:results_seal}. Moreover, in some cases, the new labels are even more accurate than the original annotation of the image (e.g. the image with a {\it car interior} label is more accurate than the {\it auto factory} label). The incorrect examples show bias problems associated with biases in some classes, where several patterns may be over-represented (e.g. {\it florist shop} appear when the network finds flower patterns).

\section{Conclusions}
\label{sec:conclusions}
Unsupervised datasets generated from web data are foreseen as key in image classification due to the volume of samples required in such problems. However, these datasets have some level of noise.  This work has addressed two types of noise present in unsupervised datasets generated from web data, namely: a) the incorrect labeling of images and b) the incomplete labeling of images. 

This work starts by presenting a system to generate a single-label dataset from web images in an unsupervised manner. The noise of this dataset is drastically reduced by performing a clustering step and a selection of clusters using anchors (therefore reducing the incorrect labeling of images). The presented system has been used to generate a dataset resulting in 250K images for 96 classes with only a 15\% of incorrectly labeled images. The percentage of correct images in the dataset presented in this paper (85\%) is notable when compared to WebVision (between 66\% and 80\% correctly labeled images).

The second contribution of this paper is aimed towards solving the incomplete labeling problem. In this context,  this paper  presents a novel system to autonomously augment the number of labels of any dataset by assigning labels to the corresponding images using the class activation maps and the uncertainty associated with each class. This system can be used to perform label augmentation in either single-label or multi-label datasets. To demonstrate the effectiveness of this system, this pipeline has been applied to the previously generated dataset and to Places365 (for reproducibility). Results have shown the addition of 9.5\% and 27\% of labels in each dataset respectively, demonstrating the applicability of this system for label augmentation (to enrich the initial dataset and describe additional visual information present on the scene). 

{\small
\bibliographystyle{ieee}
\bibliography{egbib}
}
\end{document}